\documentclass{SCIS2025}
\usepackage{multirow}

\begin{document}
\ArticleType{RESEARCH PAPER}
\Year{2025}
\Month{}
\Vol{68}
\No{}
\DOI{}
\ArtNo{000000}
\ReceiveDate{}
\ReviseDate{}
\AcceptDate{}
\OnlineDate{}
\AuthorMark{}
\AuthorCitation{}

\title{Optimal Transport Adapter Tuning for Bridging Modality Gaps in Few-Shot Remote Sensing Scene Classification}{Ji Z, Liu C, Liu J R, et al. Optimal Transport Adapter Tuning for Bridging Modality Gaps in Few-Shot Remote Sensing Scene Classification} 

\author[1,2]{Zhong JI}{}
\author[1]{Ci LIU}{}
\author[1]{Jingren LIU}{{jrl0219@tju.edu.cn}}
\author[1]{Chen TANG}{{tangchen@tju.edu.cn}}
\author[1,2]{Yanwei PANG}{}
\author[3]{Xuelong LI}{}


\address[1]{School of Electrical and Information Engineering, Tianjin Key Laboratory of Brain-inspired \\ Intelligence Technology, Tianjin University, Tianjin {\rm 300072}, China}
\address[2]{Shanghai Artificial Intelligence Laboratory, Shanghai {\rm 200232}, China}
\address[3]{Institute of Artificial Intelligence (TeleAI), China Telecom Corp Ltd, Beijing {\rm 100033}, China}

\abstract{Few-Shot Remote Sensing Scene Classification (FS-RSSC) presents the challenge of classifying remote sensing images with limited labeled samples. Existing methods typically emphasize single-modal feature learning, neglecting the potential benefits of optimizing multi-modal representations. To address this limitation, we propose a novel Optimal Transport Adapter Tuning (OTAT) framework aimed at constructing an ideal Platonic representational space through optimal transport (OT) theory. This framework seeks to harmonize rich visual information with less dense textual cues, enabling effective cross-modal information transfer and complementarity. Central to this approach is the Optimal Transport Adapter (OTA), which employs a cross-modal attention mechanism to enrich textual representations and facilitate subsequent better information interaction. By transforming the network optimization into an OT optimization problem, OTA establishes efficient pathways for balanced information exchange between modalities. Moreover, we introduce a sample-level Entropy-Aware Weighted (EAW) loss, which combines difficulty-weighted similarity scores with entropy-based regularization. This loss function provides finer control over the OT optimization process, enhancing its solvability and stability. Our framework offers a scalable and efficient solution for advancing multimodal learning in remote sensing applications. Extensive experiments on benchmark datasets demonstrate that OTAT achieves state-of-the-art performance in FS-RSSC, significantly improving the model performance and generalization.}


\maketitle

\section{Introduction}
Few-Shot Remote Sensing Scene Classification (FS-RSSC) \cite{qiu2024few} focuses on categorizing remote sensing images with limited labeled examples per class. This task is critical for applications such as land-use classification and environmental monitoring, where acquiring large-scale, high-quality labeled datasets is both costly and time-intensive. Traditional approaches to FS-RSSC leverage techniques like transfer learning \cite{pires2019convolutional}, meta-learning \cite{li2021meta}, and metric learning \cite{yuan2023few}, often using lightweight models such as Conv4, ResNet12, and ResNet18. However, these approaches struggle to fully capture the complex and diverse information in remote sensing images, leading to suboptimal performance.

\begin{figure}[t]
    \centering  
    \includegraphics[scale=0.45]{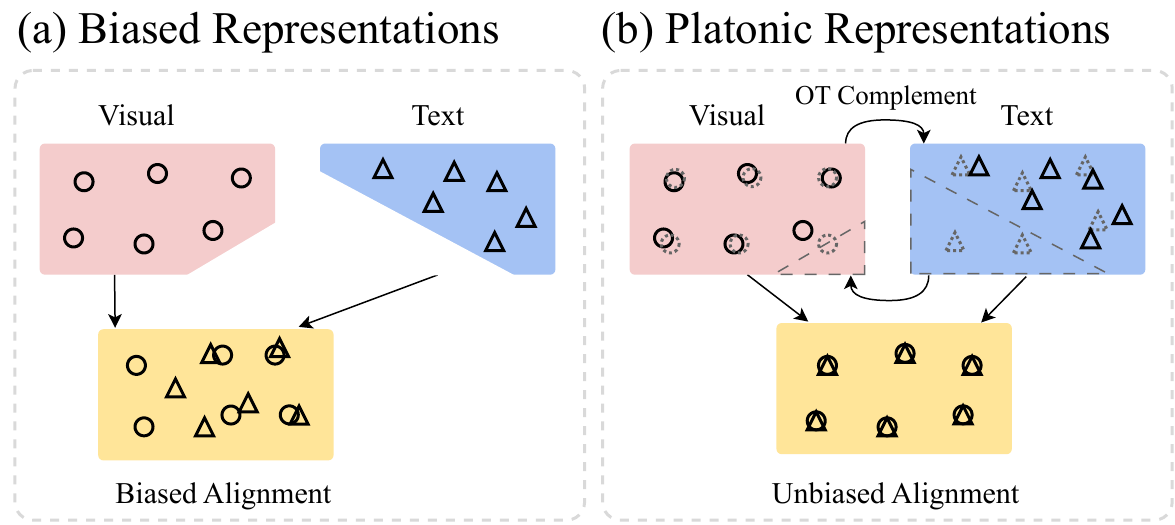}
    \caption{The illustration of our motivation. In FS-RSSC task, the disparity in information density between visual and textual modalities causes traditional optimization algorithms to disproportionately focus on the visual space, resulting in the feature space heavily biased towards visual information. The limited visual data further restricts representation learning, hindering the construction of a robust shared latent space, as illustrated in Figure~(a). In contrast, our OTAT framework leverages OT theory to guide the complementation of visual and textual information, thereby creating a unified space that approximates Platonic representations for precise alignment, as shown in Figure~(b).} 
    \label{intro_pic}
\end{figure}

With the rapid development of multi-modal models in recent years, vision-language models (VLMs), exemplified by CLIP \cite{radford2021learning}, have demonstrated exceptional generalization capabilities, especially in data-scarce scenarios. For example, the CLIP model aligns image and text embeddings through large-scale vision-language pretraining, generating robust feature representations that can integrate textual information as a complementary modality. This capability is particularly valuable for FS-RSSC, where limited labeled data can be augmented with semantic context from category names or descriptions.

To rapidly adapt VLMs for downstream tasks, building on these multi-modal foundational models, various parameter-efficient fine-tuning (PEFT) techniques, such as CoOp \cite{zhou2022learning}, Tip-Adapter \cite{zhang2022tip}, and LP++ \cite{huang2024lp++}, have emerged. However, these methods predominantly excel with natural images and face significant challenges when applied directly to remote sensing data, particularly in the few-shot setting. As illustrated in Figure~\ref{intro_pic}(a), remote sensing images possess unique characteristics, including a broad field of view and high information density, which yield semantically rich visual features. In contrast, textual information, often limited to category names, is semantically sparse. This pronounced modality gap skews the shared latent space towards the visual modality, hindering effective alignment between visual and textual features and thus limiting classification performance. The few-shot setting further exacerbates this issue, as the limited visual data restricts the ability to capture comprehensive features, making it difficult to construct a robust shared latent space. To address this challenge, drawing inspiration from the Platonic Representation Hypothesis \cite{huh2024platonicrepresentationhypothesis}, we aim to construct an unbiased shared latent space by bridging the gap between the semantically rich visual modality and the sparse textual modality. As illustrated in Figure~\ref{intro_pic}(b), our approach enhance the weaker textual modality by transferring rich semantic information from the visual modality, while simultaneously leveraging textual guidance to refine and distill essential features from the visual modality. This dynamic interplay ensures complementary and synergistic representations, enabling the construction of a shared latent space that captures the intrinsic characteristics of both modalities. By leveraging this cross-modal synergy, our approach achieves robust representation and significantly improves model performance, even under the constraints of few-shot scenarios. 

Specifically, we propose an \textbf{O}ptimal \textbf{T}ransport \textbf{A}dapter \textbf{T}uning (OTAT) framework. OTAT leverages the optimal transport (OT) theory \cite{monge1781founding} to model cross-modal complementarity and alignment, ensuring efficient and harmonious integration of information between modalities. The heart of our OTAT framework is the Optimal Transport Adapter (OTA), which utilizes a cross-modal attention mechanism to enrich textual features with corresponding visual information, and leverages the enhanced textual information to complement and augment the visual information. OTA facilitates the complementarity and synergy between semantically rich visual data and sparse textual data, thereby creating a modality-agnostic shared space that captures the conceptual essence across modalities. 

The main contributions of this paper are as follows:
\begin{itemize}
  \item We introduce the OTAT framework, which leverages the OT theory to minimize modality disparities and reduces the distributional gap between visual and textual features. This approach enables more effective cross-modal complementarity and synergy, thereby significantly enhancing FS-RSSC performance. 
  \item We propose an adapter tuning approach based on a cross-modal attention mechanism. This method reformulates the network optimization as an OT problem, enabling effective information complementarity and transfer.
  \item Furthermore, we introduce a sample-level Entropy-Aware Weighted (EAW) loss to dynamically guide the information transfer. Unlike traditional batch-level entropy regularization, EAW loss integrates difficulty-weighted similarity scores with entropy-based regularization, optimizing conditions for solving the OT problem and enhancing the precision of cross-modal alignment.
  \item We perform comprehensive ablation studies and visual analyses to validate the effectiveness of the proposed OTAT framework in bridging modality disparities and achieving unbiased cross-modal feature complementarity. Extensive experiments on four public benchmark datasets, i.e., UC Merced~\cite{yang2010bag}, NWPU-RESISC45~\cite{cheng2017remote}, AID~\cite{xia2017aid}, and WHU-RS19~\cite{sheng2012high} show that OTAT consistently outperforms the state-of-the-arts by a large margin.
\end{itemize}

\section{Related Work}
\subsection{Multimodal Alignment}
Multimodal alignment involves learning shared representations across different modalities, such as images and text, to bridge the gap between disparate data sources. Early methods rely on separate pipelines with handcrafted features and shallow alignment mechanisms, which limit scalability and generalization. Joint embedding frameworks, such as DeViSE \cite{frome2013devise} and MCB \cite{fukui2016multimodal}, introduce pre-trained encoders and cross-modal metrics but struggle in few-shot settings. Recent advancements in VLMs, exemplified by CLIP \cite{radford2021learning} and ALIGN \cite{jia2021scaling}, leverage contrastive learning on large-scale image-text datasets, achieving strong zero-shot and few-shot performance. Similarly, methods like USER \cite{zhang2024user} expand negative sample pairs in mini-batches, improving the model's ability to distinguish between positive and negative samples. Hybrid architectures, including CLIP-Adapter \cite{gao2024clip}, ModeX \cite{zhang2024modality}, and LLaVA \cite{liu2024visual}, enhance fine-grained interactions through lightweight adapters or visual encoders, further advancing state-of-the-art results. However, current multimodal alignment approaches often rely on large amounts of data and simple strategies to achieve excellent performance on natural images, often neglecting their adaptability to downstream tasks, particularly in challenging domains such as medical imaging and remote sensing. Therefore, building on the Platonic Representation Hypothesis \cite{huh2024platonicrepresentationhypothesis}, we leverage OT theory to construct the ideal representation space and demonstrates superior performance in the remote sensing domain.

\subsection{PEFT in Few-Shot Scenarios}
In recent years, VLMs have significantly advanced, becoming integral to zero-shot and few-shot applications. PEFT has emerged as a practical solution for adapting large pre-trained models to new tasks with limited labeled data. Unlike traditional full fine-tuning, which updates all parameters and risks high computational costs, PEFT employs lightweight, task-specific components while keeping most pre-trained weights frozen. Key methods include adapter-based approaches and prompt-based approaches. The former integrate task-specific layers, while the latter modify input embeddings to leverage pre-trained generalization without altering the model. Notable examples include CoOp \cite{zhou2022learning} and CoCoOp \cite{zhou2022conditional}, which optimize continuous prompt contexts for few-shot tasks, and Maple \cite{khattak2023maple}, which introduces multi-modal prompt learning to CLIP, coupling vision and language prompts across transformer blocks to enhance alignment. Adapter-based methods like CLIP-Adapter \cite{gao2024clip} adopts an additional bottleneck layer to learn new features, Tip-Adapter \cite{zhang2022tip} eliminates training with a key-value cache, and LP++ \cite{huang2024lp++} introduces an optimization method that blends image and text knowledge through learnable functions and class-specific multipliers, significantly enhance few-shot CLIP adaptation performance. Despite these advancements, few methods address the unique challenges of remote sensing in few-shot scenario, such as information imbalance across modalities and incomplete modality interaction. Our OTAT framework addresses these challenges by leveraging OT theory to enhance cross-modal information flow. This framework ensures an ideal Platonic representation space, tighter visual-textual integration, and improved generalization, all while remaining computationally efficient for remote sensing applications.

\section{Method}
\begin{figure*}[t]
	\begin{center}
        \includegraphics[width=\textwidth]{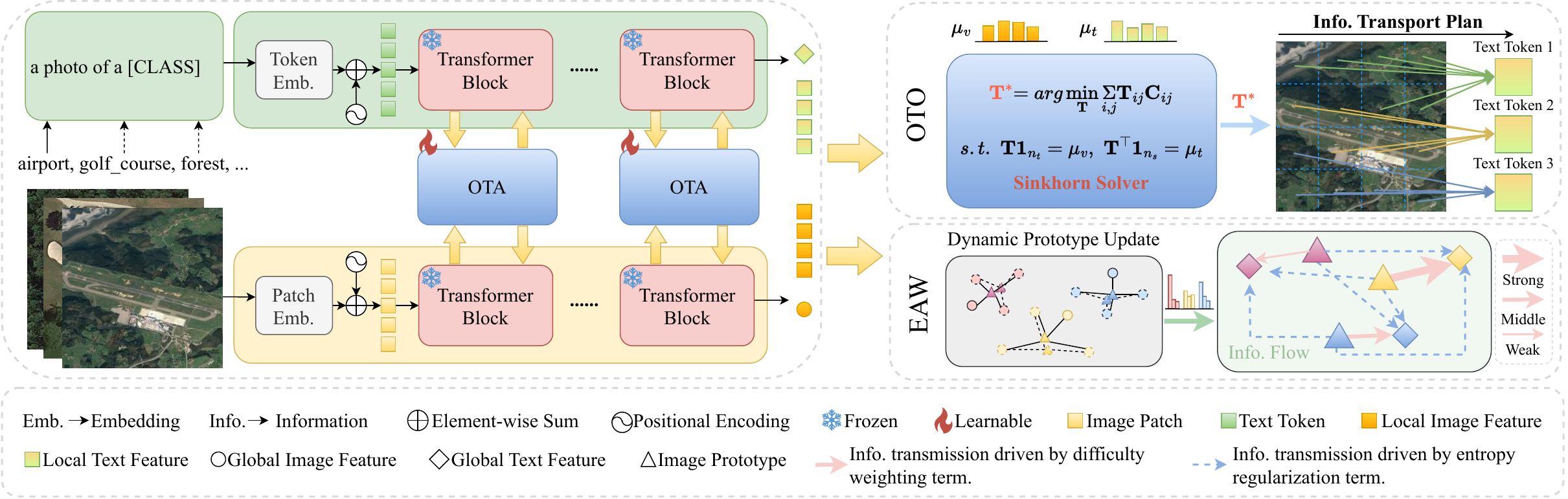}
	\end{center}
	\caption{Illustration of the proposed \textbf{O}ptimal \textbf{T}ransport \textbf{A}dapter \textbf{T}uning (OTAT) framework, comprising: (1) A frozen CLIP model and a trainable OTA structure for multimodal feature extraction; (2) OTO, which leverages visual knowledge to augment textual information and optimizes information transfer between the two modalities; and (3) EAW, integrating adaptive weight adjustment and entropy regularization to derive the optimal OT solver.}
	\label{overview}
\end{figure*}

Building upon the foundational tenets of the Platonic Representation Hypothesis, which emphasizes the criticality of unified representations in multimodal encoding, we propose the OTAT framework for FS-RSSC. As illustrated in Figure~\ref{overview}, OTAT employs OT theory to extract intricate cross-modal features, thereby minimizing the the discrepancies between modalities and achieving quintessential Platonic representations. To better optimize our OTAT framework on remote sensing datasets, we utilize the Sinkhorn-Knopp algorithm to perform optimization. In addition, we introduce the EAW loss that dynamically adjusts the alignment process by weighting image-text similarity scores according to their complexity and incorporating entropy regularization. This integrated approach transforms the initial CLIP-derived representations into their ideal Platonic counterparts, significantly enhancing performance in remote sensing applications.

Formally, let us consider a dataset \( \mathcal{D} = \{ (v_i, t_c) \}_{i=1}^N \), where \( N = C \times K \) denotes the total number of samples, encompassing \( C \) distinct classes, each with \( K \) exemplars. Here, \( v_i \) represents the \( i \)-th image datum, and \( t_c \) is the textual description corresponding to the \( c \)-th class, shared by all image samples within that class. For notational clarity, we denote a single remote sensing (RS) image as \( \mathbf{V} \in \mathbb{R}^{H \times W \times 3} \), and its associated text as \( T = \{ w_j \}_{j=1}^M \), where \( H \times W \times 3 \) specifies the spatial and channel dimensions of the image, and \( w_j \) signifies the \( j \)-th lexical unit in the text sequence.

\subsection{CLIP Representations: Initial Embeddings}
Given the remarkable generalization of CLIP model \cite{radford2021learning}, we select it as our multimodal backbone. The CLIP model integrates a visual and textual encoder, achieving robust cross-modal alignment through contrastive pre-training. However, in specialized domains such as remote sensing \cite{liu2024remoteclip}, medical imaging \cite{liu2023clip}, and embodied intelligence \cite{fan2022minedojo}, the significant differences in information density between modalities present substantial challenges to optimal performance \cite{huh2024platonicrepresentationhypothesis}.

\subsubsection{CLIP Encoders}  
The visual encoder in the CLIP model adopts a Vision Transformer (ViT) architecture \cite{dosovitskiy2020image}. The image \( \mathbf{V} \) is divided into patches and a learnable [CLS] token is prepended, generating patch embeddings. After passing through multiple transformer blocks, the [CLS] token of the final output, \( \mathbf{V}_{L_V}^{\text{cls}} \), is projected via a linear transformation \( \mathbf{W}_V \in \mathbb{R}^{D_1 \times D} \) into the shared embedding space:
\begin{equation}
    \mathbf{v} = \frac{\mathbf{V}_{L_V}^{\text{cls}} \mathbf{W}_V}{\left\| \mathbf{V}_{L_V}^{\text{cls}} \mathbf{W}_V \right\|_2}.
\end{equation}
This ensures that the global image representation, encapsulated in the [CLS] token, is aligned within the shared embedding space, while other tokens remain unaffected.

Similarly, the textual encoder processes the input text \( T \) using a Transformer \cite{vaswani2017attention}, embedding each word \( w_j \) into \( \mathbf{e}_j \in \mathbb{R}^{D_2} \) with positional embeddings. After processing through multiple transformer layers, only the [END] token is mapped into the shared space:
\begin{equation}
    \mathbf{t} = \frac{\mathbf{T}_{L_T}^{\text{end}} \mathbf{W}_T}{\left\| \mathbf{T}_{L_T}^{\text{end}} \mathbf{W}_T \right\|_2}.
\end{equation}

\subsubsection{Contrastive Pre-training}
Finally, the CLIP model is trained using a symmetric cross-entropy loss derived from the InfoNCE \cite{he2020momentum}, which promotes higher similarity for matching image-text pairs. The loss for a batch of \( N \) pairs is:
\begin{equation}
    \mathcal{L} = \frac{1}{2N} \left( \sum_{i=1}^N \mathcal{L}_{\text{image}}^i + \sum_{i=1}^N \mathcal{L}_{\text{text}}^i \right),
\end{equation}
where $\mathcal{L}_{\text{image}}^i$ and $\mathcal{L}_{\text{text}}^i$ are based on cosine similarity between image and text embeddings.

\subsubsection{Adapter-Based Fine-Tuning}
To adapt the CLIP model for the remote sensing domain, we integrate adapters into both the visual and textual encoders. Remote sensing imagery, with its high spatial resolution and specific land-use categories, differs significantly from the natural images used in CLIP's pre-training. Thus, employing PEFT techniques is essential for effective domain adaptation without full model retraining. Among PEFT approaches, adapters are notably effective\cite{cao2024bi,yuan2023parameter,wang2024rsadapter}, leading us to select them for this purpose.

For the visual encoder, the adapters are inserted in parallel with the MHSA and FFN modules of each transformer block. They are implemented with down-projection $\mathbf{W}_{\text{down}}$, non-linear activation $\sigma$, and up-projection $\mathbf{W}_{\text{up}}$, while the adapted representation of the \( l \)-th block is:
\begin{equation}
    \mathbf{V}_{l}^{\text{adapted}}=\mathbf{V}_l+\alpha \mathbf{W}_{\text{up}}\sigma \left( \mathbf{W}_{\text{down}}\mathbf{V}_{l-1} \right),
\end{equation}
where $\alpha$ is a weight coefficient and $\mathbf{V}_{l-1}$ is the \( (l-1) \)-th representation.

Similarly, the textual encoder also integrates the adapters in parallel with the MHSA and FFN modules:
\begin{equation}
    \mathbf{T}_{l}^{\text{adapted}}=\mathbf{T}_l+\alpha \mathbf{W}_{\text{up}}\sigma \left( \mathbf{W}_{\text{down}}\mathbf{T}_{l-1} \right).
\end{equation}

Despite the parameter efficiency of this construction, it does not fully address the challenges posed by the imbalance in semantic density between remote sensing images and text. These characteristics constrain CLIP's cross-modal alignment capabilities, even with fine-tuned adapters. To surmount these limitations, we employ OT theory to reengineer the structure and optimization of the adapters, generating embeddings that more closely approximate the ideal Platonic representations.

\subsection{Platonic Representations: Optimal Transport Adapters}
To approximate the ideal Platonic representations—abstract, modality-agnostic embeddings that capture the essence of concepts across different modalities, we introduce an innovative architecture grounded in OT theory. Our objective is to minimize discrepancies between visual and textual feature distributions by efficiently transporting information across modalities, thereby creating a unified embedding space. At the core of our framework is the Optimal Transport Adapters (OTA) structure, which leverages the cross-modal attention mechanism to enrich textual features with visual information. By formulating the alignment between modalities as an OT problem, we aim to find the most efficient mapping of visual features onto textual ones, minimizing the ``cost" of this transfer and ensuring harmonious integration.

Specifically, given an image \( \mathbf{V} \), we extract its local patch features \( \left\{ \mathbf{V}_{L_V}^{i} \right\}_{i=1}^{L_1} \) from the final transformer block of the CLIP visual encoder. These features are then mapped into a shared embedding space using a linear transformation \( \mathbf{W}_V \), resulting in the visual feature map \( \mathbf{V}_{\text{local}} = \left[ \mathbf{v}^1, \mathbf{v}^2, \dots, \mathbf{v}^{L_1} \right]^\top \in \mathbb{R}^{L_1 \times D} \). Similarly, we process the textual data by mapping the outputs of the text encoder into the same embedding space, yielding the textual feature map \( \mathbf{T}_{\text{local}} = \left[ \mathbf{t}^{\text{start}}, \mathbf{t}^1, \mathbf{t}^2, \dots, \mathbf{t}^{L_2}, \mathbf{t}^{\text{end}} \right]^\top \in \mathbb{R}^{(L_2 + 2) \times D} \).

\subsubsection{OTA Structure}
\begin{figure}[t]
    \centering  
    \includegraphics[scale=0.5]{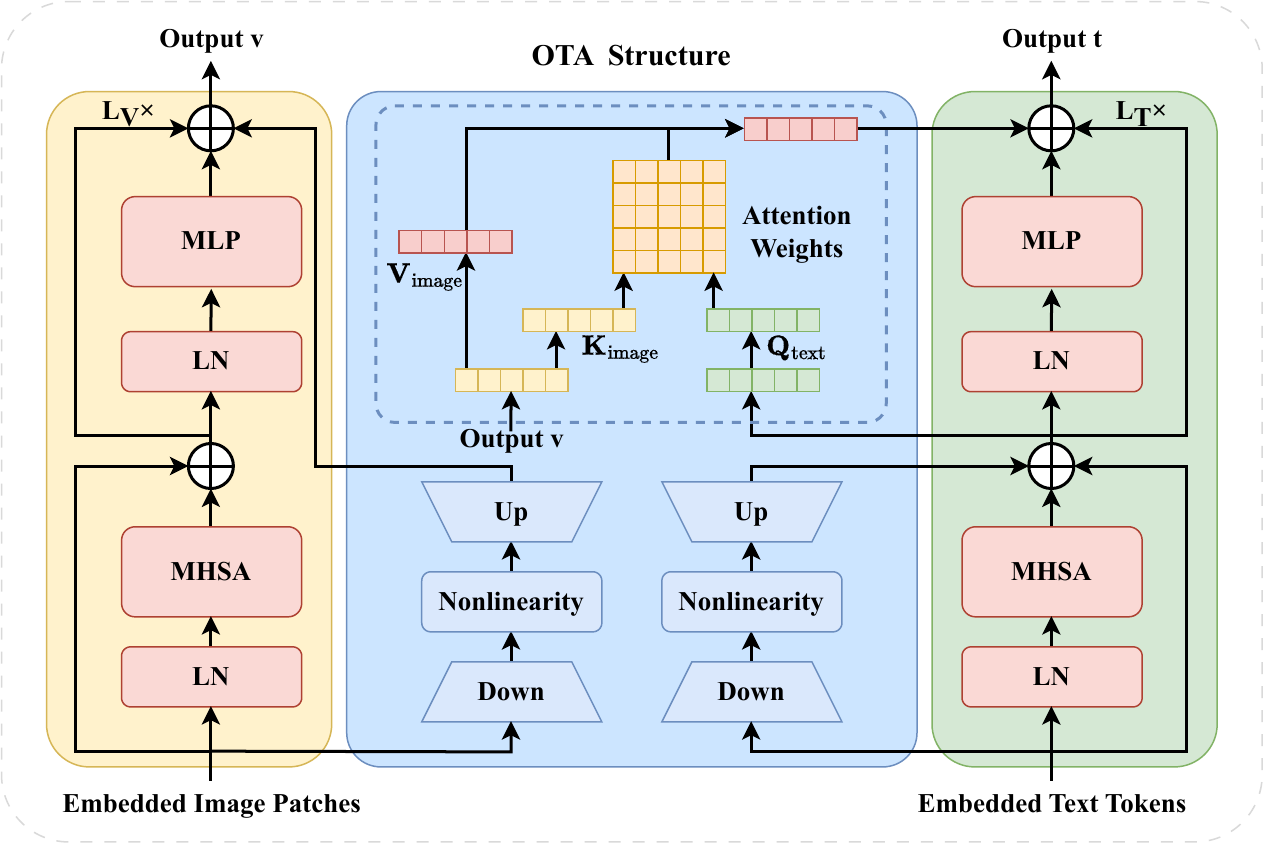}
    \caption{Implementation details of the OTA structure. In the image encoder, adapter layers are placed parallel to both MHSA and FFN blocks. In the text encoder, adapter layers are positioned parallel to MHSA blocks, with a cross-modal attention mechanism added parallel to the FFN block.} 
    \label{ot-structure}
\end{figure}

Due to the significant information density differences between image and text modalities in FS-RSSC, direct information transfer fails to perfectly achieve the desired alignment. To overcome this, we propose the OTA structure, which leverages cross-modal attention mechanism to enrich the textual information with image features. As shown in Figure~\ref{ot-structure}, we enhance the basic adapter structure by placing the adapters parallel to the MHSA module, and integrating a cross-modal attention mechanism alongside the FFN module. This design not only mitigates the textual information deficiency but also serves as a medium for aligning cross-modal information, preparing the model for further optimization through dynamic, adaptive information transfer driven by the OT plan.

First, in each transformer block, the input features undergo the following transformation:
\begin{equation}
    \mathbf{T}_{l}^{\text{res}} = \mathbf{T}_{l}^{\text{att}} + \alpha \mathbf{W}_{\text{up}} \sigma \left( \mathbf{W}_{\text{down}} \mathbf{T}_{l-1} \right),
\end{equation}
where \( \alpha \) is a scaling factor, \( \mathbf{W}_{\text{up}} \) and \( \mathbf{W}_{\text{down}} \) are the adapter parameters, and \( \sigma \) is the activation function.

To accomplish modal information transfer, we utilize cross-modal attention operations between the textual \( \mathbf{T}_{l}^{\text{res}} \) and visual \( \mathbf{v}_i \) from multiple images within the same class. These features are projected into a common query-key-value space:
\begin{equation}
\mathbf{Q}_{l}^{\text{text}}=\mathbf{W}_{Q}^{\text{text}}\mathbf{T}_{l}^{\text{res}},\ \mathbf{K}_{l}^{\text{image}_i}=\mathbf{W}_{K}^{\text{image}}\mathbf{v}_i,\ \mathbf{V}_{l}^{\text{image}_i}=\mathbf{W}_{V}^{\text{image}}\mathbf{v}_i,\ 
\end{equation}
where \( \mathbf{W}_{Q}^{\text{text}}, \mathbf{W}_{K}^{\text{image}}, \mathbf{W}_{V}^{\text{image}} \in \mathbb{R}^{D \times D} \) represent trainable projection matrices that facilitate mapping the input features into a unified latent space for effective cross-modal interaction.

The attention weights, which are conducive to subsequent transmission, are then computed as:
\begin{equation}
    \mathbf{W}_{l}^{\text{cmam}_i} = \text{softmax} \left( \frac{\mathbf{Q}_{l}^{\text{text}} (\mathbf{K}_{l}^{\text{image}_i})^\top}{\sqrt{D}} \right).
\end{equation}

As a medium for information transfer, this structure is not only optimized for the subsequent OT plan optimization, it also determine how textual features attend to visual features, effectively transferring information from the visual modality. The enriched textual features are then aggregated:
\begin{equation}
    \mathbf{T}_{l}^{\text{cmam}} = \mathbf{T}_{l}^{\text{res}} + \gamma \cdot \frac{1}{K} \sum_{i=1}^{K} \mathbf{W}_{l}^{\text{cmam}_i} \mathbf{V}_{l}^{\text{image}_i},
\end{equation}
where \( \gamma \) is a scaling factor, and \( K \) is the number of visual samples from the same category.

The enriched features are processed via the FFN module:
\begin{equation}
    \mathbf{T}_{l}^{\text{enhanced}} = \text{FFN} \left( \text{LN} \left( \mathbf{T}_{l}^{\text{res}} \right) \right) + \beta \mathbf{T}_{l}^{\text{cmam}},
\end{equation}
where \( \beta \) balances the contribution of the enriched features, and \( \text{LN} \) denotes the layer normalization.

By integrating the OT structure within adapters, we ensure efficient and theoretically grounded feature transport between modalities. This structure creates a unified, modality-agnostic space, enriching textual information with relevant visual information, leading to features that closely approximate ideal Platonic forms and enhancing the cross-modal interactions.

\subsubsection{OTA Optimization} 
To facilitate smoother interactions between modalities, we propose an optimization approach based on OT solver. We consider the visual and textual features as discrete probability distributions:
\begin{equation}
    P_v = \sum_{i=1}^{L_1} a_i \delta_{\mathbf{v}^i}, \quad P_t = \sum_{j=1}^{L_2 + 2} b_j \delta_{\mathbf{t}^j},
\end{equation}
where \(\delta_{\mathbf{x}}\) is the Dirac delta function at \(\mathbf{x}\), and \(a_i\) and \(b_j\) are uniform weights, i.e., \(a_i = \frac{1}{L_1}\) and \(b_j = \frac{1}{L_2 + 2}\).

The cost function \( c(\mathbf{v}^i, \mathbf{t}^j) \) measures the dissimilarity between \(\mathbf{v}^i\) and \(\mathbf{t}^j\) via cosine distance:
\begin{equation}
    c(\mathbf{v}^i, \mathbf{t}^j) = 1 - \frac{\langle \mathbf{v}^i, \mathbf{t}^j \rangle}{\|\mathbf{v}^i\| \|\mathbf{t}^j\|}.
\end{equation}

With the cost fixed, the optimal transport problem seeks a transport plan \(\mathbf{T} = [T_{ij}] \in \mathbb{R}_+^{L_1 \times (L_2 + 2)}\) that minimizes the total transportation cost from \(P_v\) to \(P_t\):
\begin{equation}
    W_c(P_v, P_t) = \min_{\mathbf{T} \in \Pi(P_v, P_t)} \sum_{i=1}^{L_1} \sum_{j=1}^{L_2 + 2} T_{ij} \, c(\mathbf{v}^i, \mathbf{t}^j),
\end{equation}
subject to the marginal constraints:
\begin{equation}
    \small
    \Pi(P_v, P_t) = \left\{ \mathbf{T} \in \mathbb{R}_+^{L_1 \times (L_2 + 2)} \ \bigg| \ \sum_{j=1}^{L_2 + 2} T_{ij} = a_i, \ \sum_{i=1}^{L_1} T_{ij} = b_j \right\}.
\end{equation}

To ensure stability and efficient computation, we introduce the the Kantorovich relaxation\cite{kantorovich2006translocation}:
\begin{equation}
    W_c^\lambda(P_v, P_t) = \min_{\mathbf{T} \in \Pi(P_v, P_t)} \sum_{i,j} T_{ij} \, c(\mathbf{v}^i, \mathbf{t}^j) - \lambda H(\mathbf{T}),
\label{ot_distance}
\end{equation}
where \( H(\mathbf{T}) = -\sum_{i,j} T_{ij} \log T_{ij} \) is the entropy of the transport plan, and \(\lambda > 0\) is the regularization parameter.

We solve this problem using the Sinkhorn algorithm\cite{sinkhorn1967diagonal}, which iteratively approximates the entropy-regularized OT plan while satisfying the marginal constraints.

First, we define the cost matrix \(\mathbf{C} \in \mathbb{R}^{L_1 \times (L_2 + 2)}\):
\begin{equation}
    \mathbf{C}_{ij} = c(\mathbf{v}^i, \mathbf{t}^j) = 1 - \frac{\langle \mathbf{v}^i, \mathbf{t}^j \rangle}{\|\mathbf{v}^i\| \|\mathbf{t}^j\|}.
\end{equation}

Next, we define the kernel matrix \(\mathbf{K}\):
\begin{equation}
    \mathbf{K}_{ij} = \exp(-\lambda \mathbf{C}_{ij}).
\end{equation}

We then initialize vectors \(\mathbf{p} = 1_{L_1}\) and \(\mathbf{q} = 1_{L_2 + 2}\), and iteratively update them as follows:
\begin{align} \label{ota_iteration}
    \mathbf{p} &\gets \frac{a}{\mathbf{K} \mathbf{q}}, \quad \mathbf{q} \gets \frac{b}{\mathbf{K}^\top \mathbf{p}},
\end{align}
until convergence. The convergence criterion is based on the Frobenius norm, where we stop when the difference between successive transport plans is below a threshold $\varepsilon _1$, i.e., $\lVert \mathbf{T}^{(k+1)} - \mathbf{T}^{(k)} \rVert_F < \varepsilon _1$. 

Upon convergence, the optimal transport plan \(\mathbf{T}^*\) is given by:
\begin{equation}
    \mathbf{T}^* = \text{diag}(\mathbf{p}) \mathbf{K} \text{diag}(\mathbf{q}).
\end{equation}

Finally, the entropy-regularized optimal transport distance is computed as:
\begin{equation}
    W_c^\lambda(P_v, P_t) = \langle \mathbf{T}^*, \mathbf{C} \rangle = \sum_{i=1}^{L_1} \sum_{j=1}^{L_2 + 2} \mathbf{T}_{ij}^* \mathbf{C}_{ij}.
\end{equation}

Using the Sinkhorn algorithm, we efficiently and stably solve the entropy-regularized OT problem, facilitating robust information exchange between visual and textual features.

After computing the entropy-regularized OT distance \(W_c^\lambda(P_v, P_t)\), we define the probability of a match between an image \(\mathbf{v}_i\) and text \(\mathbf{t}_j\) as:
\begin{equation}
    p_{\text{OT}}(y=j|\mathbf{v}_i) = \frac{\exp\left((1 - W_c^\lambda(P_i, P_j))/\tau\right)}{\sum_{c=1}^{C}\exp\left((1 - W_c^\lambda(P_i, P_c))/\tau\right)},
\end{equation}
where \(W_c^\lambda(P_i, P_j)\) is the OT distance between the \(i\)-th image and the text of the \(j\)-th category.

To enhance cross-modal information transport, with the fixed optimal transport plan \(\mathbf{T}^*\), we apply the loss function:
\begin{equation}
    \mathcal{L}_{\text{OTA}} = -\frac{1}{N}\sum_{i \in N, j \in C} y_{i,j} \log p_{\text{OT}}(y=j|\mathbf{v}_i).
    \label{ot_loss}
\end{equation}

This approach ensures precise, high-quality cross-modal alignment, guiding the model towards ideal, modality-independent Platonic representations.

\subsection{Regularizing Pathway: Entropy-Aware Weighted Loss}
With the aid of (\ref{ot_loss}), the CLIP model is initially optimized to approximate Platonic representations. However, this optimization assumes uniform treatment of all samples within a batch, which may be suboptimal in FS-RSSC due to significant sample variability. To alleviate this, we introduce the Entropy-Aware Weighted (EAW) loss, which dynamically adjusts the optimization process based on sample complexity. Unlike the batch-level entropy regularization in (\ref{ot_distance}), EAW loss incorporates difficulty-weighted similarity and entropy regularization at the sample level, offering more granular control over information flow.

\subsubsection{Dynamic Prototype Update}
We compute the class prototypes using a dynamic update strategy. The class prototypes $\mathbf{p}_c$ are initialized to zero and updated using the average of the normalized visual embeddings for each class $c$ as:
\begin{equation}
    \mathbf{p}_c=\frac{1}{n_c}\sum_{j=1}^{n_c}{\frac{\mathbf{v}_{\left( j \right)}}{\lVert \mathbf{v}_{\left( j \right)} \rVert}},
\end{equation}
where $n_c$ is the number of samples in class $c$.

The update rule incorporates a weighting factor $\mu$, which increases over time, stabilizing near 1:
\begin{equation}
    \mathbf{p}_{c}^{\text{updated}}=\mu \mathbf{p}_{c}^{\text{old}}+\left( 1-\mu \right) \mathbf{p}_{c}^{\text{new}}.
\end{equation}

\subsubsection{EAW loss} The key contribution of EAW loss is its sample-level entropy regularization, which promotes a balanced similarity distribution across all pairs, ensuring smoother information flow between image and text modalities. The entropy regularization term is defined as:
\begin{equation}
    r=-\frac{1}{C}\sum_{c,j\in C}{p_{c,j}\log \left( p_{c,j}+\varepsilon _2 \right)},
\end{equation}
where
\begin{equation}
    p_{c,j}=\frac{\exp \left( \text{sim}\left( \mathbf{p}_{c}^{\text{updated}},\mathbf{t}_j \right) /\tau \right)}{\sum_{j=1}^C{\exp}\left( \text{sim}\left( \mathbf{p}_{c}^{\text{updated}},\mathbf{t}_j \right) /\tau \right)},
\end{equation}
and $\varepsilon _2$ ensures numerical stability.

In addition to entropy regularization, EAW loss incorporates a difficulty-weighting mechanism to further refine the alignment of challenging samples. The difficulty-weighted similarity loss adjusts the focus based on the hardest examples: 
\begin{equation}
    \mathcal{L}_{\text{difficulty}}=-\frac{1}{C}\sum_{c,j\in C}{k_cp_{c,j}},
\end{equation}
where \( k_c = 1 - \max_j p_{c,j} \). This ensures that more challenging image-text pairs receive greater attention.

The total EAW loss is formulated as:
\begin{equation}
    \mathcal{L}_{\text{EAW}}=\mathcal{L}_{\text{difficulty}}-\zeta r,
\end{equation}
where $\zeta$ balances the similarity distribution uniformity and the focus on challenging samples.

By prioritizing sample-level entropy regularization, EAW loss maintains a balanced similarity distribution while adapting to the complexity of individual samples. The difficulty-weighting mechanism complements this by emphasizing challenging examples. Together, these components enhance the effectiveness of the OTA structure, refining CLIP-derived representations to better approximate ideal Platonic forms and improving the model's performance in remote sensing tasks.

\subsection{Total Loss}
In addition to the OTA loss and EAW loss, we incorporate a cosine similarity loss, \(\mathcal{L}_{\cos}\), to maintain the base classification performance in FS-RSSC, which is defined as:
\begin{equation}
    \small
    \mathcal{L}_{\cos} = -\frac{1}{N} \sum_{i \in N, j \in C} y_{i,j} \log \frac{\exp \left( \text{sim}(\mathbf{v}_i, \mathbf{t}_i) / \tau \right)}{\sum_{j=1}^C \exp \left( \text{sim}(\mathbf{v}_i, \mathbf{t}_j) / \tau \right)}.
\end{equation}

These three losses collectively refine the CLIP-derived features into an ideal Platonic features. Specifically, the OTA loss enhances textual information by leveraging visual knowledge and optimizes the information transfer between modalities. The EAW loss introduces entropy regularization and adaptive weight adjustment to address sample importance imbalances, ensuring more effective and balanced feature alignment. The cosine similarity loss $\mathcal{L}_{\cos}$ ensures the maintenance of base performance. Formally, the total loss is expressed as:
\begin{equation}
    \mathcal{L}_{train}=\mathcal{L}_{\cos}+\xi\mathcal{L}_{\text{OTA}}+\nu\mathcal{L}_{\text{EAW}},
\end{equation}
where $\xi$, $\nu$ control the contributions of each component. 

\section{Experiments Results and Analysis}
In this section, we conduct a detailed analysis and experiments to evaluate the performance of our OTAT framework in the FS-RSSC scenarios. We also perform thorough ablation studies, comparative experiments, and visualizations to validate the effectiveness of each component and our motivations.

\subsection{Experimental Setup}
\subsubsection{Datasets}
In the experimental section, we evaluate our proposed OTAT framework on four prominent remote sensing datasets: UC Merced~\cite{yang2010bag}, NWPU-RESISC45~\cite{cheng2017remote}, AID~\cite{xia2017aid}, and WHU-RS19~\cite{sheng2012high}. Inspired by CoOp~\cite{zhou2022learning}, we divide each dataset into training, validation, and testing sets, comprising 30\%, 20\%, and 50\% of the data per class.
\begin{itemize}
    \item \textbf{UC Merced~\cite{yang2010bag}}: Released by University of California, Merced. It includes 2100 images in 21 classes, and the pixel size per image is $256 \times 256$.
    \item \textbf{NWPU-RESISC45~\cite{cheng2017remote}}: Originating from Northwestern Polytechnical University, it consists of 45 scene categories, with 700 images per category, each of size $256 \times 256$ pixels.
    \item \textbf{AID~\cite{xia2017aid}}: Co-released by Wuhan University and Huazhong University of Science and Technology, this dataset features 10,000 images across 30 categories, with resolutions of $600 \times 600$ pixels.
    \item \textbf{WHU-RS19~\cite{sheng2012high}}: Compiled by Wuhan University utilizing Google Earth, this dataset comprises 1,005 images, each $600 \times 600$ pixels, covering 19 categories with no fewer than 50 images per category.
\end{itemize}

\subsubsection{Implementation Details}
In this study, all experimental procedures are executed on a single NVIDIA RTX 4090 GPU, employing the pretrained CLIP ViT-B/16 as the foundational model. To meet the input size requirements, remote sensing images are resized to 224 × 224 pixels. Accordingly, the textual prompt template adopts the form ``a photo of a [CLS]". For the hyper-parameters in our OTAT framework, the down-projection dimension of the adapter is set to 8, the weight coefficient $\alpha$ is designated as 1.0, and the regularization strength $\lambda$ is configured to 10. The iterative process of (\ref{ota_iteration}) for the OTA optimization is limited to a maximum of 100 cycles. For dynamic prototype updating, the initial value of $\mu$ is 0.5, which is incremented by 0.02 at each step. Training configurations include an initial learning rate of 0.001, leveraging the AdamW optimizer, and the network is trained with a batch size of 64 across 50 epochs. To facilitate adaptive learning rate adjustments, a cosine annealing schedule is implemented throughout the training period.

\subsection{Comparisons with State-of-the-Art Approaches}
To elucidate the advantages of our proposed framework, FS-RSSC experiments are conducted across four distinct datasets, facilitating a direct comparison with state-of-the-art methodologies. These benchmark evaluations encompass both CNN-based approaches, such as SCL-MLNet\cite{li2021scl}, TSC\cite{zeng2022task}, MPCL-Net\cite{ma2023multi}, TDNet\cite{wang2023tdnet}, TeAW\cite{cheng2023teaw}, and ACL-Net\cite{xu2024attention}, as well as ViT-based approaches, including zero-shot CLIP\cite{radford2021learning}, full fine-tuning CLIP\cite{radford2021learning}, Linear Probe\cite{radford2021learning}, Tip-Adapter-F\cite{zhang2022tip}, CLIP-Adapter\cite{gao2024clip}, LP++\cite{huang2024lp++}, CoOp\cite{zhou2022learning}, CoCoOp\cite{zhou2022conditional}, MaPLe\cite{khattak2023maple}, RemoteCLIP\cite{liu2024remoteclip}. This comprehensive assessment ensures a rigorous validation of our OTAT framework's efficacy and robustness in diverse few-shot scenarios.

To ensure a fair comparison with CNN-based approaches, we adopt the meta-learning paradigm for 5-way 1-shot and 5-way 5-shot experiments, following the data partitioning from \cite{li2020dla} and \cite{li2021scl}. For the ViT-based approaches, we perform full-way classification across 1- to 32-shot scenarios. Additionally, we categorize the ViT-based approaches into zero-shot, full fine-tuning, adapter-based, and prompt-based approaches to underscore the distinct advantages of our proposed framework. The comparative results in Table~\ref{UCandWHU} and Table~\ref{NWPUandAID} demonstrate the superior performance of ours approach across four datasets.

\begin{table*}[t]
\centering
\setlength{\tabcolsep}{3pt} 
\caption{Few-shot classification accuracy (\%) comparisons on the UC MERCED and WHU-RS19 datasets.}
\resizebox{\linewidth}{!}{
    \begin{tabular}{cccp{0.9cm}<{\centering}p{0.9cm}<{\centering}p{0.9cm}<{\centering}p{0.9cm}<{\centering}p{0.9cm}<{\centering}p{0.9cm}<{\centering}p{0.9cm}<{\centering}|p{0.9cm}<{\centering}p{0.9cm}<{\centering}p{0.9cm}<{\centering}p{0.9cm}<{\centering}p{0.9cm}<{\centering}p{0.9cm}<{\centering}p{0.9cm}<{\centering}}
    \toprule
    \multirow{2}{*}{Approach} & \multirow{2}{*}{Backbone} & \multirow{2}{*}{Setting} & \multicolumn{7}{c}{UC MERCED} & \multicolumn{7}{c}{WHU-RS19} \\
    &       &     & 1-shot & 2-shot & 4-shot & 5-shot & 8-shot & 16-shot & 32-shot & 1-shot & 2-shot & 4-shot & 5-shot & 8-shot & 16-shot & 32-shot \\
    \midrule
    SCL-MLNet$_{\rm_{TGRS'22}}$\cite{li2021scl}  & \multirow{7}{*}{ResNet12}  & \multirow{7}{*}{5-way} & 51.37 & - & - & 68.09 & - & - & - & 63.36 & - & - & 77.62 & - & - & - \\
    TSC$_{\rm_{ISPRS'22}}$\cite{zeng2022task}   & &  &  55.11 & - & - & 69.20 & - & - & - & 70.99 & - & - & 82.18 & - & - & - \\
    MPCL-Net$_{\rm_{TGRS'23}}$\cite{ma2023multi}  & &  &  56.46 & - & - & 76.57 & - & - & - & 61.84 & - & - & 80.34 & - & - & - \\
    TDNet$_{\rm_{TGRS'23}}$\cite{wang2023tdnet}   & & &   - & - & - & - & - & - & - & 64.24 & - & - & 84.15 & - & - & - \\
    TeAW$_{\rm_{ICASSP'23}}$\cite{cheng2023teaw}  & & &  56.94 & - & - & \underline{77.50} & - & - & - & - & - & - & - & - & - & - \\
    ACL-Net$_{\rm_{TGRS'24}}$\cite{xu2024attention}   & & & \underline{59.74} & - & - & 74.89 & - & - & - & \underline{78.30} & - & - & \underline{90.43} & - & - & - \\
    \textbf{OTAT (Ours)} & CLIP & & \textbf{96.90} & - & - & \textbf{99.11} & - & - & - & \textbf{95.79} & - & - & \textbf{98.53} & - & - & - \\
    \midrule
    Zero-shot$^{*}_{\rm_{ICML'21}}$\cite{radford2021learning}     & \multirow{11}{*}{CLIP}  & \multirow{11}{*}{full-way}  & \multicolumn{7}{c|}{65.81} & \multicolumn{7}{c}{71.40} \\
    Full fine-tuning$^{*}_{\rm_{ICML'21}}$\cite{radford2021learning}   & & & \underline{82.86} & 83.43 & \underline{92.29} & \underline{92.38} & \underline{92.95} & \underline{96.38} & \underline{96.67} & \underline{93.19} & \underline{94.94} & 96.50 & \underline{98.64} & \underline{98.64} & \underline{99.22} & \underline{99.42} \\
    Linear Probe$^{*}_{\rm_{ICML'21}}$\cite{radford2021learning}      & & & 63.32 & 76.29 & 85.53 & 87.52 & 90.32 & 93.02 & 93.51 & 79.98 & 90.19 & 94.30 & 96.07 & 97.30 & 97.65 & 97.67 \\
    CoOp$^{*}_{\rm_{IJCV'22}}$\cite{zhou2022learning}   & &  & 76.10 & 83.52 & 86.38 & 88.76 & 91.24 & 95.14 & 96.57 & 87.94 & 91.44 & 95.14 & 96.30 & 96.11 & 97.86 & 97.28 \\
    CoCoOp$^{*}_{\rm_{CVPR'22}}$\cite{zhou2022conditional}  &  & & 74.10 & 78.19 & 82.10 & 85.14 & 87.52 & 89.05 & 91.81 & 87.94 & 90.08 & 94.16 & 93.39 & 95.72 & 96.89 & 95.91 \\
    MaPLe$^{*}_{\rm_{CVPR'23}}$\cite{khattak2023maple}   & &  & 64.83 & 74.51 & 74.70 & 75.11 & 79.05 & 83.87 & 85.46 & 80.67 & 82.36 & 86.45 & 87.03 & 91.96 & 93.32 & 94.94 \\
    Tip-Adapter-F$^{*}_{\rm_{ECCV'22}}$\cite{zhang2022tip}   & & & 78.67 & 82.10 & 85.90 & 86.19 & 88.67 & 92.95 & 95.05 & 92.02 & 92.83 & 96.50 & 96.69 & 96.50 & 97.47 & 98.64 \\
    CLIP-Adapter$^{*}_{\rm_{IJCV'24}}$\cite{gao2024clip}   & & & 69.71 & 71.05 & 75.43 & 77.52 & 81.24 & 83.33 & 90.19 & 76.46 & 78.21 & 85.02 & 83.46 & 91.44 & 94.75 & 94.36 \\
    RemoteCLIP$^{\dag}_{\rm_{TGRS'24}}$\cite{liu2024remoteclip} & &  & - &- & - & - & - & - & - & 40.88 & - & 60.88 & - & 80.75 & 89.45 & 93.38 \\
    LP++$^{*}_{\rm_{CVPR'24}}$\cite{huang2024lp++}   & &  & 78.76 & \underline{84.53} & 88.32 & 88.58 & 91.21 & 93.36 & 93.63 & 90.58 & 94.38 & \underline{96.58} & 96.83 & 97.70 & 97.80 & 98.13 \\
    \textbf{OTAT (Ours)}    & &  & \textbf{84.67} & \textbf{88.10} & \textbf{92.67} & \textbf{95.62} & \textbf{95.90} & \textbf{97.14} & \textbf{97.81} & \textbf{94.55} & \textbf{95.13} & \textbf{97.28} & \textbf{98.83} & \textbf{99.03} & \textbf{99.42} & \textbf{99.61} \\
    \bottomrule
    \multicolumn{17}{l}{\footnotesize * indicates results reproduced by us. Methods using CLIP adopt ViT-B/16, except those with \dag, which use ViT-B/32.}
    \end{tabular}
}
\label{UCandWHU}
\end{table*}

\begin{table*}[t]
\centering
\setlength{\tabcolsep}{3pt} 
\caption{Few-shot classification accuracy (\%) comparisons on the NWPU-RESISC45 and AID datasets.}
\resizebox{\linewidth}{!}{
\begin{tabular}{cccp{0.9cm}<{\centering}p{0.9cm}<{\centering}p{0.9cm}<{\centering}p{0.9cm}<{\centering}p{0.9cm}<{\centering}p{0.9cm}<{\centering}p{0.9cm}<{\centering}|p{0.9cm}<{\centering}p{0.9cm}<{\centering}p{0.9cm}<{\centering}p{0.9cm}<{\centering}p{0.9cm}<{\centering}p{0.9cm}<{\centering}p{0.9cm}<{\centering}}
    \toprule
    \multirow{2}{*}{Approach} & \multirow{2}{*}{Backbone} & \multirow{2}{*}{Setting} & \multicolumn{7}{c}{NWPU-RESISC45} & \multicolumn{7}{c}{AID} \\
    &       &     & 1-shot & 2-shot & 4-shot & 5-shot & 8-shot & 16-shot & 32-shot & 1-shot & 2-shot & 4-shot & 5-shot & 8-shot & 16-shot & 32-shot \\
\midrule
SCL-MLNet$_{\rm_{TGRS'22}}$\cite{li2021scl}  & \multirow{7}{*}{ResNet12}  & \multirow{7}{*}{5-way} & 62.21 & - & - & 80.86 & - & - & - & 59.46 & - & - & 76.31 & - & - & - \\
TSC$_{\rm_{ISPRS'22}}$\cite{zeng2022task}     &  &  & 73.26 & - & - & 84.62 & - & - & - & 68.21 & - & - & 81.72 & - & - & - \\
MPCL-Net$_{\rm_{TGRS'23}}$\cite{ma2023multi}  &  & & 55.94 & - & - & 76.24 & - & - & - & 60.61 & - & - & 76.78 & - & - & - \\
TDNet$_{\rm_{TGRS'23}}$\cite{wang2023tdnet}  & &  & 65.85 & - & - & 82.16 & - & - & - & 67.48 & - & - & 80.56 & - & - & - \\
TeAW$_{\rm_{ICASSP'23}}$\cite{cheng2023teaw}  &  &  & 70.23 & - & - & 85.57 & - & - & - & 70.35 & - & - & 84.62 & - & - & - \\
ACL-Net$_{\rm_{TGRS'24}}$\cite{xu2024attention}  &  &  & \underline{76.13} & - & - & \underline{86.54} & - & - & - & \underline{70.86} & - & - & \underline{85.37} & - & - & - \\
\textbf{OTAT (Ours)} & CLIP & & \textbf{93.28} & - & - & \textbf{95.96} & - & - & - & \textbf{96.42} & - & - & \textbf{98.56} & - & - & - \\
\midrule
Zero-shot$^{*}_{\rm_{ICML'21}}$\cite{radford2021learning}     & \multirow{11}{*}{CLIP}    & \multirow{11}{*}{full-way}  & \multicolumn{7}{c|}{57.00} & \multicolumn{7}{c}{57.20} \\
Full fine-tuning$^{*}_{\rm_{ICML'21}}$\cite{radford2021learning}     &  &  & 70.36 & \underline{78.25} & \textbf{82.36} & \underline{83.84} & \textbf{85.86} & \textbf{87.42} & \underline{89.14} & 77.22 & 82.46 & 87.68 & \textbf{90.94} & \underline{91.64} & \textbf{94.44} & \underline{95.36} \\
Linear Probe$^{*}_{\rm_{ICML'21}}$\cite{radford2021learning}     &  &  & 52.45 & 64.72 & 75.31 & 77.35 & 81.79 & 85.46 & 88.22 & 64.66 & 78.38 & 85.96 & 87.25 & 91.05 & 93.78 & 94.84 \\
CoOp$^{*}_{\rm_{IJCV'22}}$\cite{zhou2022learning}     &  &  & 63.31 & 70.11 & 76.30 & 77.68 & 80.29 & 84.62 & 87.63 & 72.12 & 82.14 & 82.90 & 86.66 & 87.34 & 90.72 & 92.70 \\
CoCoOp$^{*}_{\rm_{CVPR'22}}$\cite{zhou2022conditional}     &  &  & 69.88 & 71.53 & 74.36 & 75.78 & 79.15 & 80.72 & 82.10 & 71.54 & 73.50 & 79.84 & 84.06 & 84.58 & 87.68 & 89.68 \\
MaPLe$^{*}_{\rm_{CVPR'23}}$\cite{khattak2023maple}  & &  & 57.10 & 59.43 & 63.93 & 63.48 & 68.51 & 72.58 & 76.16 & 60.17 & 65.65 & 64.86 & 69.91 & 74.90 & 80.83 & 84.59 \\
Tip-Adapter-F$^{*}_{\rm_{ECCV'22}}$\cite{zhang2022tip}     &  &  & 67.33 & 71.30 & 76.13 & 78.24 & 82.98 & 86.48 & 88.43 & \underline{78.86} & 82.78 & 87.22 & 86.76 & 90.30 & 93.32 & 95.08 \\
CLIP-Adapter$^{*}_{\rm_{IJCV'24}}$\cite{gao2024clip}    &  &  & 63.37 & 66.74 & 68.69 & 70.99 & 74.65 & 79.89 & 84.24 & 64.02 & 62.66 & 69.56 & 70.42 & 74.30 & 87.60 & 92.30 \\
RemoteCLIP$^{\dag}_{\rm_{TGRS'24}}$\cite{liu2024remoteclip} & &  & 42.22 &- & 60.86 & - & 70.97 & 75.88 & 81.83 & 36.98 & - & 65.62 & - & 75.71 & 81.07 & 86.72 \\
LP++$^{*}_{\rm_{CVPR'24}}$\cite{huang2024lp++}    &  &  & \underline{71.59} & 76.07 & 80.37 & 81.80 & 84.00 & 86.19 & 88.17 & 77.68 & \underline{84.76} & \underline{88.33} & 89.31 & 91.57 & 93.44 & 94.32 \\
\textbf{OTAT (Ours)}      &  &  & \textbf{72.02} & \textbf{78.81} & \underline{81.70} & \textbf{84.03} & \underline{84.76} & \underline{87.37} & \textbf{89.61} & \textbf{79.12} & \textbf{85.50} & \textbf{88.62} & \underline{90.12} & \textbf{92.50} & \underline{94.22} & \textbf{95.70} \\
\bottomrule
\multicolumn{17}{l}{\footnotesize * indicates results reproduced by us. Methods using CLIP adopt ViT-B/16, except those with \dag, which use ViT-B/32.}
\end{tabular}
}
\label{NWPUandAID}
\end{table*}

\begin{figure*}[t]
      \centering
      \captionsetup[subfigure]{labelfont=small}
      \subfloat[\scriptsize Baseline]{
           \includegraphics[scale=0.3]{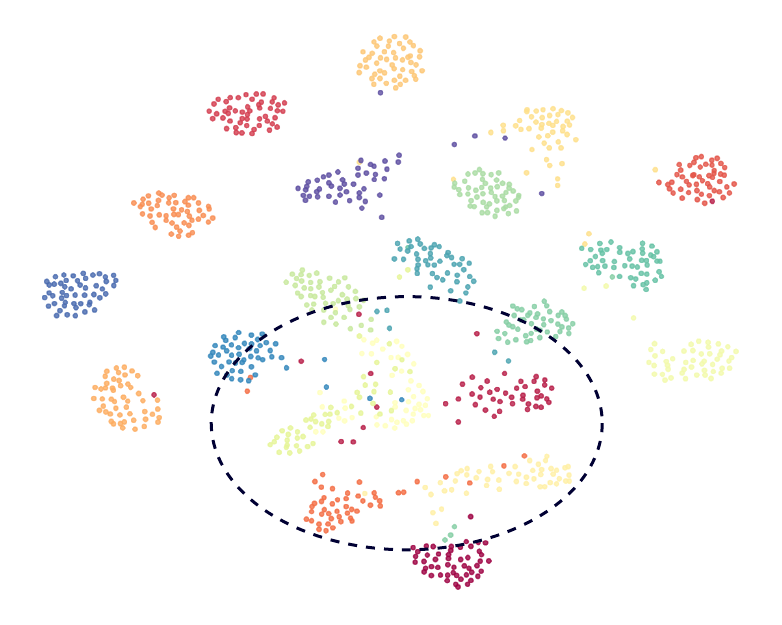}\label{vis1}
      }\hspace*{-5pt}
      \subfloat[\scriptsize OTO]{
           \includegraphics[scale=0.3]{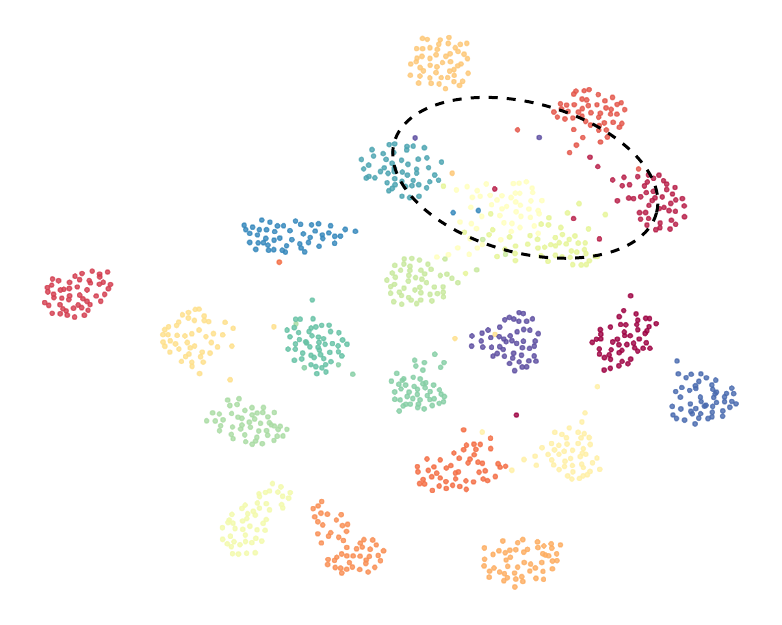}\label{vis2}
      }\hspace*{-5pt}
      \subfloat[\scriptsize OTA + OTO]{
           \includegraphics[scale=0.3]{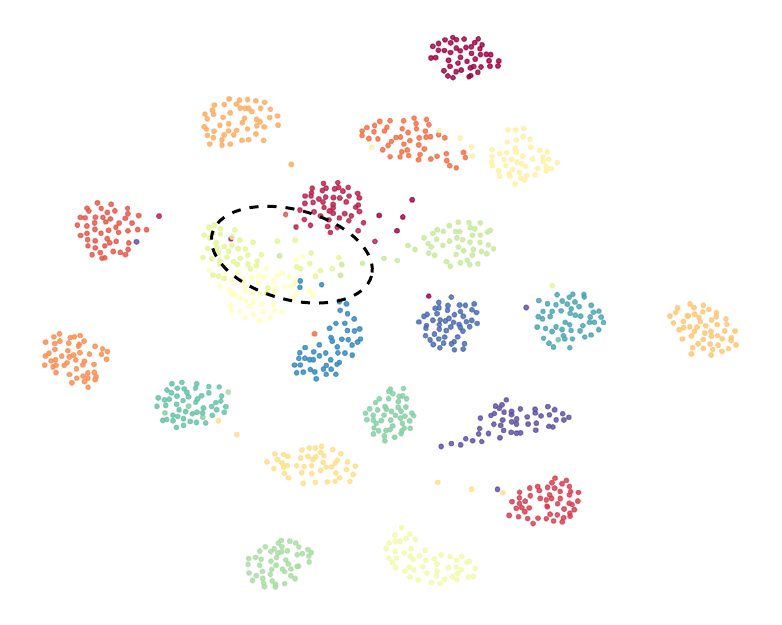}\label{vis3}
      }\hspace*{-5pt}
      \subfloat[\scriptsize OTA + OTO + EAW]{
           \includegraphics[scale=0.3]{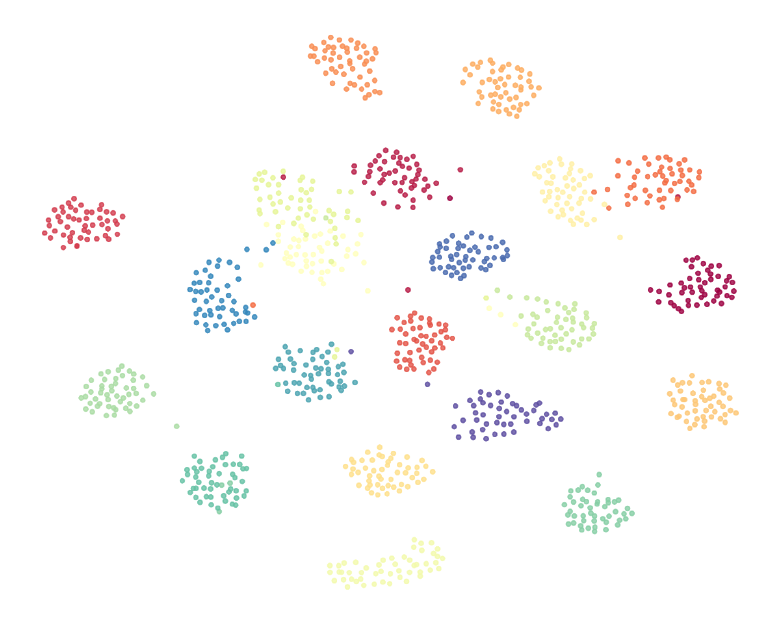}\label{vis4}
      }\hspace*{-5pt}
      \caption{The UMAP visualization of generated features under different configurations: (a) Baseline, where the original adapters are optimized with standard cross-entropy loss, (b) OT Optimization (OTO), applying our OT optimization to the original adapters, (c) OTA + OTO, integrating OT optimization into our OTA structure, and (d) OTA + OTO + EAW, further incorporating EAW loss. Dashed lines highlight regions with noticeable class mixing.}\label{UMAP}
\end{figure*}

As shown in Table~\ref{UCandWHU}, our OTAT framework significantly improves performance on smaller datasets, outperforming CNN-based approaches by 37.16\% and 21.61\% in the 1-shot and 5-shot settings on the UC MERCED dataset, and by 17.49\% and 8.10\% on the WHU-RS19 dataset. This highlights the superior representation learning capacity of large-scale pre-trained models. Furthermore, our framework consistently outperforms adapter-based and prompt-based approaches, with improvements of 1.24\%-6.86\% on UC MERCED dataset and 0.7\%-2.02\% on WHU-RS19 dataset relative to the best-performing alternatives. This underscores its effectiveness in leveraging OT theory to capture cross-modal feature correlations, facilitating precise information complementarity.

Notably, our OTAT framework outperforms full fine-tuning, showing improvements of 0.38\% to 4.67\% on the UC Merced dataset and 0.19\% to 1.36\% on the WHU-RS19 dataset. This is particularly significant, as full fine-tuning is often considered the upper bound of performance. These results validate that the theoretically optimal Platonic representation, as approximated by ours, is superior to empirical full fine-tuning. This finding provides a promising direction for the future development of multi-modal learning.

In addition, as shown in Table~\ref{NWPUandAID}, our OTAT framework remains effective in more complex scenarios. On the NWPU-RESISC45 dataset, it outperforms CNN-based approaches by 17.15\% and 9.42\% in 1-shot and 5-shot settings, respectively, and on the AID dataset, it improves by 25.56\% and 13.19\%, demonstrating its robustness in high-complexity, multi-class classification tasks. Moreover, our framework consistently outperforms other PEFT approaches across both datasets and frequently surpasses full fine-tuning. This highlights OTAT's ability to provide an efficient and precise information transfer pathway, ensuring effective modality balance even in challenging scenarios. However, while OTAT excels in low-shot settings, its performance slightly trails full fine-tuning as the number of training samples increases, emphasizing its advantage in low-resource scenarios. By leveraging an efficient information transfer mechanism, OTAT enables rapid adaptation with limited data, providing a practical alternative to full fine-tuning. 

Overall, these excellent results validate our motivation for generating ideal Platonic representations and the importance of the various components based on OT theory. These components consistently produce superior performance across various few-shot scenarios.

\subsection{Ablation Study}
\subsubsection{Representation Learning across Various Module Combinations}
To evaluate the contribution of each component in guiding the CLIP model toward an approximate ideal Platonic representation, we visualize the learned visual features and employ Mutual Nearest Neighbors (MNN) as a metric to assess cross-modal alignment performance. Experiments are conducted under four configurations: (1) a baseline configuration where the original adapters are inserted into each transformer block of the frozen CLIP model and optimized with standard cross-entropy loss, (2) applying our OT optimization (OTO) to optimize the original adapters, (3) applying OTO to our proposed OTA structure, and (4) incorporating the EAW loss into the optimization process.

As shown in Figure~\ref{UMAP}, the baseline demonstrates significant inter-class overlap, with multiple feature points clustered into other classes. Introducing OT theory effectively mitigates these issues by redistributing the weights between image and text features through information transfer, enhancing intra-class consistency and reducing inter-class confusion. Building on this foundation, adding the OTA structure significantly enhances the uniformity of feature distributions. This improvement arises from the cooperative optimization of both modalities, where text features are enriched with complementary image information. Finally, integrating EAW loss refines the feature learning process even further, nearly eliminating misclassifications and improving separability between challenging categories.

\begin{figure}[!t]
  \centering
  \begin{minipage}[c]{0.48\textwidth}
  \centering
  \includegraphics[width=0.95\textwidth]{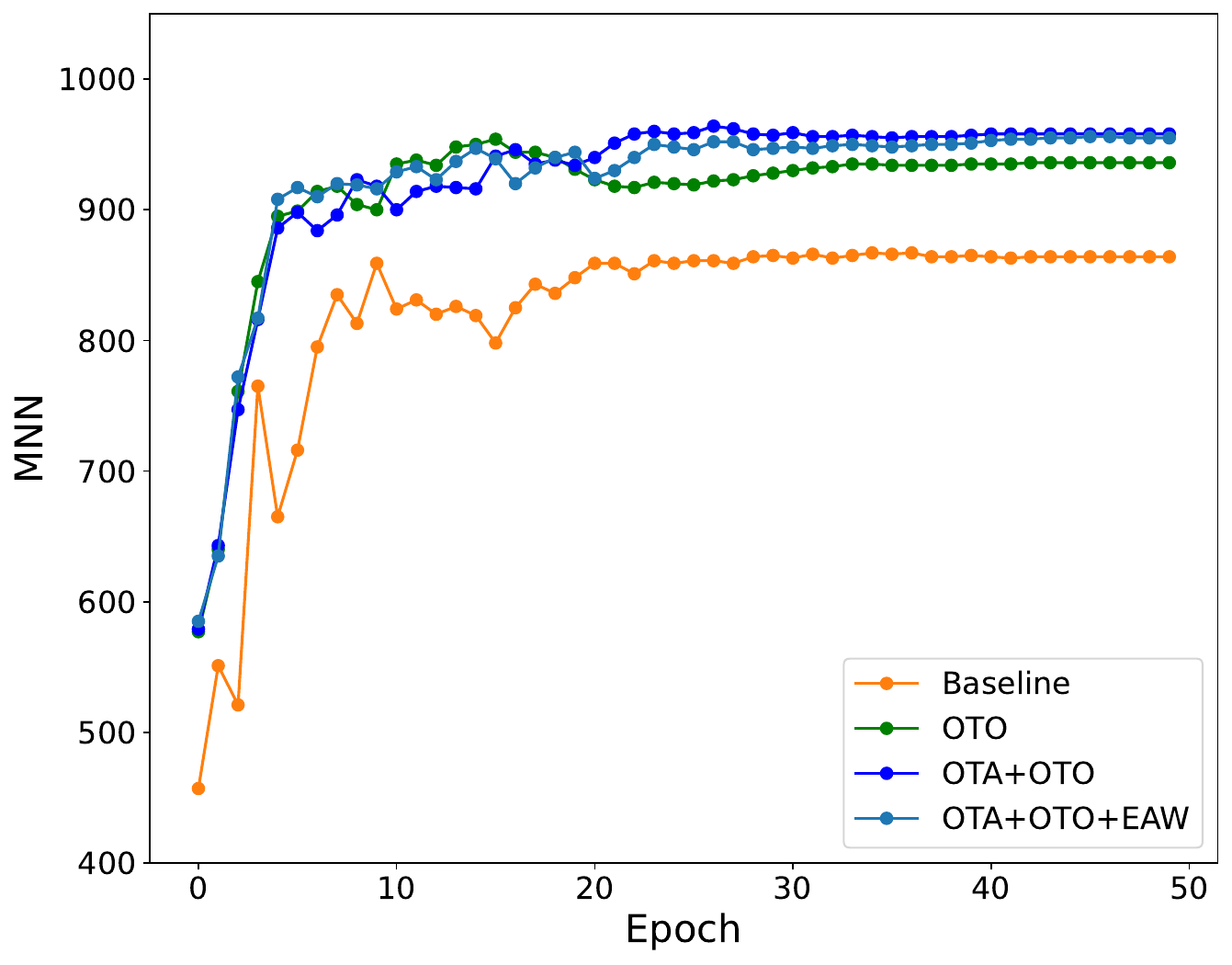}
  \end{minipage}
  \hspace{0.02\textwidth}
  \begin{minipage}[c]{0.48\textwidth}
  \centering
  \includegraphics[width=0.95\textwidth]{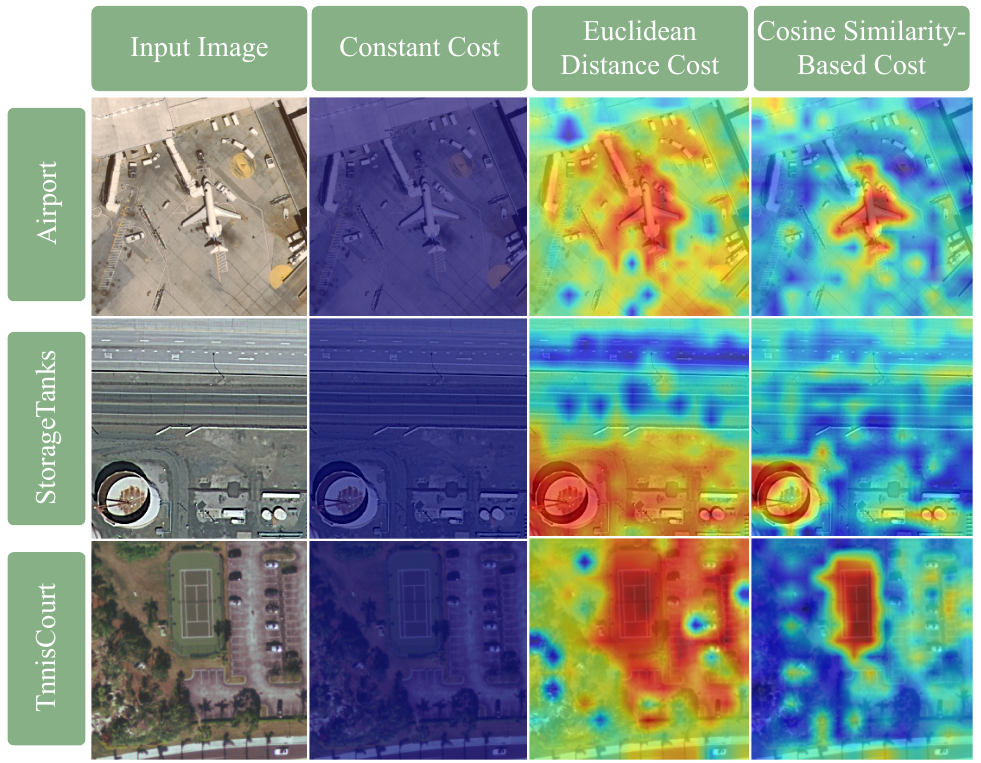}
  \end{minipage}\\[3mm]
  \begin{minipage}[t]{0.48\textwidth}
  \centering
  \caption{Impact of module combinations on cross-modal alignment. The MNN metric is used to measure alignment over training.}
  \label{mnn}
  \end{minipage}
  \hspace{0.02\textwidth}
  \begin{minipage}[t]{0.48\textwidth}
  \centering
  \caption{Ablation study on the optimal transport cost function. Warm colors indicate regions with smaller cumulative transport distances to text tokens, highlighting their greater contribution to textual information.}
  \label{OTheatmap}
  \end{minipage}
  \end{figure}

Additionally, to demonstrate that we have achieved the construction of theoretically optimal Platonic representations, we provide the change curve of modal alignment performance in Figure~\ref{mnn}. With the introduction of OTO, modal alignment is significantly enhanced, indicating that text modality information is effectively enriched and consistent features are learned through image-text information transmission. Building on this, the inclusion of the OTA structure facilitates smoother alignment, with the alignment performance stabilizing at a higher level, highlighting the importance of modality information interaction in optimal transport methods. Finally, incorporating EAW loss as a sample-level regularization term ensures the necessary conditions for solving the entropy-regularized OT problem. Our framework achieves high-quality modal alignment while creating the most ideal Platonic space.

These results validate the effectiveness of each component in advancing cross-modal representation learning, culminating in a unified and aligned Platonic representation space with superior performance.

\subsubsection{Classification Performance with Different Module Combinations}
To further evaluate the effectiveness of each component, we conduct an ablation study under the full-way 5-shot setting. The results, shown in Table \ref{component-ablation}, highlight the significant contributions of each component. Introducing OTO leads to notable performance improvements of 0.19\%-2.51\% across four datasets, demonstrating its crucial role in enhancing information transfer and compensating for weaker modality. Compared to using OTO alone, adding the OTA structure further improves performance by 0.33\% - 2.24\%, as the OTA structure minimizes transmission losses and facilitates more robust multi-modal interactions. Finally, incorporating EAW loss results in the best performance, highlighting its effectiveness as a sample-level regularizer. Collectively, these components not only individually enhance the model's performance but synergistically contribute to a more robust and efficient representation learning framework, validating their complementary roles in creating ideal Platonic feature representations.

\begin{table}[t]
  \centering
  \caption{Ablation study (\%) on component effectiveness under the full-way 5-shot setting.}
  \resizebox{0.7\linewidth}{!}{
  \begin{tabular}{ccccccc}
  \toprule
   OTO &  OTA structure &  $\mathcal{L}_{\text{EAW}}$ & UC MERCED & WHU-RS19 & NWPU-RESISC45 & AID \\
  \midrule
                      &                      &                      & 91.52   & 97.67   &   80.59   &    85.84 \\
   \checkmark          &                      &                      & 93.90    & 97.86      &83.10     &    87.54                  \\
   \checkmark          & \checkmark           &                      & 94.38  &  98.64   &  83.43    &89.78                       \\
   \checkmark          &\checkmark                     & \checkmark           & 95.62   & 98.83       &84.03    &       90.12      \\
  \bottomrule
  \end{tabular}
  }
  \label{component-ablation}
\end{table}

\begin{table}[t]
  \centering
  \caption{Ablation study on optimal transport cost function in the full-way 5-shot setting of the UC Merced dataset.}
  \resizebox{0.5\linewidth}{!}{
  \begin{tabular}{ccc}
  \toprule
    Cost function &  Accuracy(\%) &  Train time(s) \\
  \midrule
    Constant cost &       91.52        &    71.43     \\
   Euclidean distance cost &      90.29     &    78.69       \\
   Cosine similarity-based cost       & 93.90      &  75.26 \\
  \bottomrule
  \end{tabular}
  }
  \label{costmatrix-ablation}
\end{table}

\subsubsection{Optimal Transport Cost Function}
To investigate the impact of the optimal transport cost function \( c(\mathbf{v}^i, \mathbf{t}^j) \) on information transmission, we design three types of cost functions: constant cost (serving as a baseline control), Euclidean distance cost, and cosine similarity-based cost (used in our framework). As shown in Table~\ref{costmatrix-ablation}, the Euclidean distance cost results in a noticeable drop in accuracy, highlighting its limitations in capturing nuanced relationships between features. In contrast, the cosine similarity-based cost effectively models the angular relationships between feature vectors, which is crucial for aligning multimodal representations. This superiority can be attributed to the ability of cosine similarity to emphasize relative directions over magnitudes, enabling a more precise transportation plan that enhances model performance. Furthermore, the marginal increase in training time for the cosine similarity-based approach is well-justified by its substantial performance gains, making it the most robust choice.

To investigate how OT theory directs information transfer between image patches and textual tokens, we visualize the cumulative OT distance on the original images as heatmaps, with the value for the $i$-th patch computed as $1-\sum_{j=1}^{L_2+2}{\mathbf{T}_{ij}^{*}\mathbf{C}_{ij}}$. As shown in Figure~\ref{OTheatmap}, the results highlight the effectiveness of the cosine similarity-based cost function, which more precisely focused on class-specific key features, such as the fuselage and wings of airplanes or the distinct layouts of tennis courts. This indicates a progressive enhancement in the alignment between visual and textual modalities. Compared to alternative cost functions, the cosine similarity-based approach dynamically refines feature mapping, enabling more precise and interpretable information transfer and complementarity.

\begin{figure}[t]
    \centering  
    \includegraphics[scale=0.4]{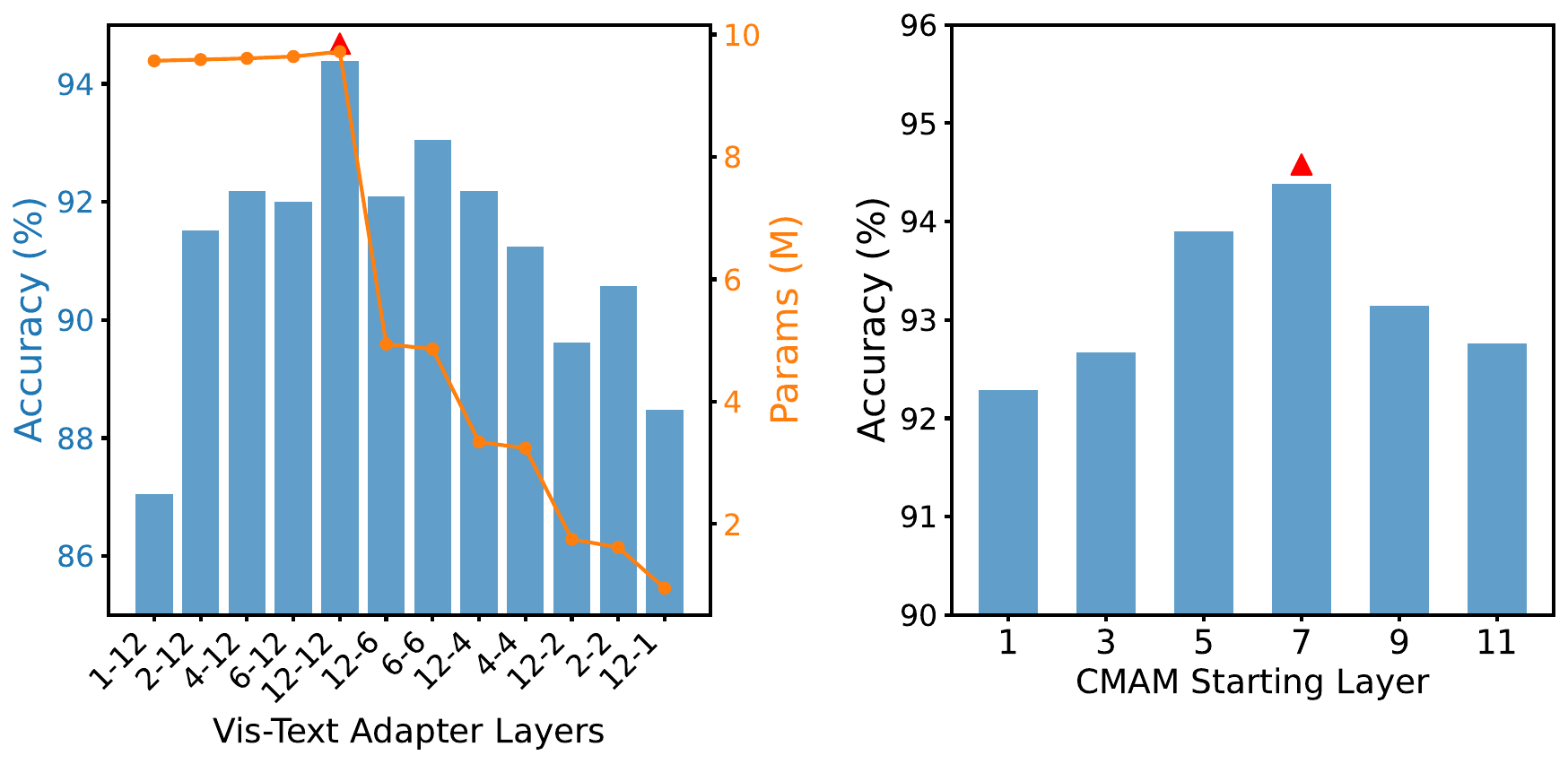}
    \caption{Ablation study on adapter insertion strategy. On the left, the number of inserted adapters is varied, with ``1V-12T" indicating 1 adapter in the visual encoder and 12 adapters in the text encoder. The curve represents the parameter count, while the histogram illustrates the classification accuracy. On the right, the starting layer for the cross-modal attention mechanism is varied to assess its impact on classification performance.} 
    \label{Adapter-Insertion-Strategy}
\end{figure}

\begin{table}[t]
    \centering
    \caption{Impact of $\mathcal{L}_{\text{OTA}}$ weight coefficient $\xi$ in the full-way 5-shot setting of the UC Merced dataset.}
    \resizebox{0.6\linewidth}{!}{
    \begin{tabular}{ccccccc}
    \toprule
    Meteric & Value 1 & Value 2 & Value 3 & Value 4 & Value 5 & Value 6 \\
    \midrule
    weight & 0.01 & 0.1 & 1.0 & 2.0 & 5.0 & 10.0    \\
    Accuracy(\%) & 90.29 & 92.29 & 93.90 & 93.14 & 93.52 & 90.00 \\
    \bottomrule
    \end{tabular}
    }
    \label{ot_weight}
\end{table} 

\begin{table}[t]
  \centering
  \caption{Impact of $\mathcal{L}_{\text{EAW}}$ weight coefficient $\nu$ in the full-way 5-shot setting of the UC Merced dataset.}
  \resizebox{0.6\linewidth}{!}{
  \begin{tabular}{ccccccc}
  \toprule
  Meteric & Value 1 & Value 2 & Value 3 & Value 4 & Value 5 & Value 6 \\
  \midrule
   weight & 0.01 & 0.04 & 0.08 & 0.12 & 0.16 & 0.2 \\
   Accuracy(\%) & 93.33 & 93.52 & 95.62 & 94.29 & 92.38 & 93.14 \\
  \bottomrule
  \end{tabular}
  }
  \label{EAW loss_weight}
\end{table} 

\subsubsection{Adapter Insertion Strategy}
To further demonstrate the superiority of our OTA component, we conduct an analysis of adapter insertion strategies, focusing on two aspects: the number of inserted adapters and the starting layer of the cross-modal attention mechanism (CMAM). These strategies target different dimensions of adapter integration and collectively influence the model’s ability to achieve optimal performance. 

As shown in the left panel of Figure~\ref{Adapter-Insertion-Strategy}, the collaborative design of vis-adapters and text-adapters significantly impacts model performance and parameter efficiency. With limited textual information, increasing the number of text-adapters enhances text representation and overall performance. Conversely, adding vis-adapters enhances accuracy with abundant text-adapters, emphasizing the importance of visual information. A balanced adapter structure, such as 6 vis-adapters and 6 text-adapters, achieves 93.05\% accuracy with 4.86M parameters, offering a strong trade-off. These findings demonstrate that flexible adapter configurations can optimize performance for specific scenarios while effectively managing resource consumption. As shown in the right panel of Figure~\ref{Adapter-Insertion-Strategy}, the starting position of the CMAM significantly influences performance. Starting at layer 7 achieves the best accuracy, balancing foundational feature extraction and higher-level semantic interaction. Starting too early disrupts low-level feature learning, while starting too late limits the refinement of cross-modal alignment. Intermediate layers, such as layer 5, also perform well, demonstrating their ability to balance both aspects effectively. These results underscore the importance of strategically selecting the CMAM starting layer to optimize performance.

\subsection{Weighting Coefficients Analysis}
\subsubsection{Impact of OTA loss weight coefficient}
Table \ref{ot_weight} highlight the critical impact of the \(\mathcal{L}_{\text{OTA}}\) weight $\xi$ on optimizing cross-modal classification accuracy. Accuracy increases from 90.29\% to 92.29\% in the low-weight region and peaks at 93.90\% at $\xi=1.0$, demonstrating the power of OT theory in guiding effective information complementarity between modalities. Beyond this optimal point, higher weights (e.g., 2.0–5.0) lead to moderate declines, and excessive emphasis (e.g., weight 10.0) disrupts this harmony, degrading accuracy to 90.00\%. This may be due to excessive weights causing the network to focus solely on information complementarity and alignment between different modalities, thereby neglecting the classification performance.

\subsubsection{Impact of EAW loss weight coefficient}
\(\mathcal{L}_{\text{EAW}}\) ensures smooth cross-modal information flow by dynamically adjusting sample contributions. As shown in Table \ref{EAW loss_weight}, the accuracy peaks at 95.62\% when $\nu = 0.08$. Smaller coefficients fail to emphasize challenging samples adequately, while larger coefficients over-regularize the model, disrupting global feature learning. The optimal value strikes a balance between emphasizing challenging samples and maintaining robust global representations. This balance enables the model to refine cross-modal similarity distributions, achieving a more refined representation space and the highest accuracy.

\begin{table}[t]
  \centering
  \caption{Impact of entropy weight coefficient $\zeta$ in the full-way 5-shot setting of the UC Merced dataset.}
  \resizebox{0.6\linewidth}{!}{
  \begin{tabular}{cccccccc}
  \toprule
  Meteric & Value 1 & Value 2 & Value 3 & Value 4 & Value 5 & Value 6 \\
  \midrule
   weight & 0.01 & 0.05 & 0.1 & 0.15 & 0.2 & 0.5 \\
   Accuracy(\%) & 93.71 & 93.05 & 95.62 & 94.10 & 93.52 & 92.19 \\
  \bottomrule
  \end{tabular}
  }
  \label{entropy_weight}
\end{table}

\begin{table}[t]
  \centering
  \caption{Comparison with state-of-the-art methods for cross-dataset generalization.}
  \resizebox{0.6\linewidth}{!}{
  \begin{tabular}{ccccc}
  \toprule
  \multirow{2}{*}{Approach} &  Source      & \multicolumn{3}{c}{Target}   \\
    &  NWPU-RESISC45  & UC MERCED & WHU-RS19  & AID  \\
  \midrule
  CoOp$^{*}$\cite{zhou2022learning}  & 77.68 & 78.57 & 81.71 & 58.42 \\
  CoCoOp$^{*}$\cite{zhou2022conditional}  & 75.78 & 80.57 & \underline{84.05} & \textbf{68.10} \\
  CLIP-Adapter$^{*}$\cite{gao2024clip} & 70.99 & 76.57 & 80.35 & 65.54 \\
  APPLeNet$^{*}$\cite{singha2023applenet} & \underline{76.29} & \underline{80.86} & 70.04 & 65.78  \\
  \textbf{OTAT(Ours)} & \textbf{84.03} & \textbf{84.48} & \textbf{87.35} & \underline{68.02} \\
  \bottomrule
  \multicolumn{5}{l}{\footnotesize * indicates results reproduced by us.}
  \end{tabular}
  }
  \label{CrossDataset}
\end{table}

\subsubsection{Impact of entropy weight coefficient}
To evaluate the effect of entropy weight $\zeta$ on model performance within $\mathcal{L}_{\text{EAW}}$, we fix \(\mathcal{L}_{\text{EAW}}\) weight to 0.08 and vary $\zeta$. As shown in Table \ref{entropy_weight}, accuracy improves to a peak of 95.62\% at $\zeta = 0.1$. Further increases to 0.2 and 0.5 reduce accuracy to 93.52\% and 92.19\%, respectively, indicating the adverse effects of excessive weighting. These findings indicate that an entropy weight of 0.1 is optimal for the solvability of the Sinkhorn algorithm under Kantorovich relaxation conditions.

\subsection{Cross-Dataset Generalization}
Additionally, to further demonstrate the strong generalization capabilities of the final generated Platonic representations, we present the results of cross-dataset generalization experiments in Table \ref{CrossDataset}, comparing the performance of OTAT with several state-of-the-art methods. Using NWPU-RESISC45 as the source dataset, the models are evaluated on UC MERCED, WHU-RS19, and AID as target datasets. The results show that OTAT consistently achieves the best or second-best performance across all target datasets. Notably, OTAT achieves 87.35\% on WHU-RS19 dataset and 84.48\% on UC MERCED dataset, significantly outperforming other methods. While CoCoOp and APPLeNet exhibit competitive performance on specific datasets, their overall stability and generalization capabilities are inferior to our OTAT framework. Similarly, CoOp and CLIP-Adapter show limitations across different target datasets. The superior performance of our framework can be attributed to its innovative integration of OT-based information transfer and entropy-aware weighted loss design. These components effectively enhance cross-modal alignment and generate Platonic representations, resulting in improved generalization for remote sensing datasets.

\section{Conclusion}
In this paper, we propose the OTAT framework to address modality asymmetry challenges in multimodal learning, particularly for remote sensing applications. By leveraging OT theory, OTAT effectively bridges the modality gap, generating modality-agnostic Platonic representations. The framework integrates three key components: an adapter structure for enriched feature interaction, OT-based optimization for efficient cross-modal information transfer, and entropy-aware regularization for sample-level balance. Extensive experiments across benchmark datasets demonstrate OTAT's superiority over state-of-the-art methods, achieving improved few-shot classification accuracy and robust cross-dataset generalization. These results highlight OTAT's potential as a scalable and computationally efficient solution for multimodal representation learning, offering a promising direction for multi-modal remote sensing domains.

\Acknowledgements{This work was supported by the National Natural Science Foundation of China (Grant No.62176178). }


\bibliographystyle{scis}
\bibliography{OTAT}

\end{document}